\documentclass[a4paper, 10 pt, conference]{hsmr} 
\usepackage[utf8]{inputenc}
\IEEEoverridecommandlockouts                       
\usepackage{mathtools}
\usepackage{geometry}
\usepackage[labelsep=space]{caption}
\usepackage{newtxmath,newtxtext}
\usepackage{authblk}
\usepackage{pgfplots}
\usepackage{graphicx}
\usepackage[colorlinks,urlcolor=blue]{hyperref} 
\usepackage{newtxtext,newtxmath}
\usepackage{mhchem}
\usepackage{fancyhdr}
\usepackage{marginnote}

\pgfplotsset{compat=newest} 
 
\title{\LARGE \bf
Deep Imitation Learning for Automated Drop-In Gamma Probe Manipulation\vspace{-12pt}} 

\author{\LARGE Kaizhong Deng}
\author{\LARGE Baoru Huang} 
\author{\LARGE Daniel S. Elson\vspace{-10pt}} 

\affil{\large\textit{The Hamlyn Centre for Robotic Surgery, Imperial College London, London, UK}\\ \large\textit{k.deng21@imperial.ac.uk}\vspace{-12pt}}

\begin{document}

\maketitle
\thispagestyle{empty}
\pagestyle{empty}
\thispagestyle{fancy}
\section*{INTRODUCTION}
Prostate cancer is one of the most common cancers in the UK, and Robotic-Assisted Surgery (RAS) has become a common method for prostate cancer surgery. Identification of cancer, extended lymph node dissection and sentinel lymph node (SLN) biopsy are current or desirable aspects of prostate cancer surgery, and SLN can provide tailored treatment with accuracy comparable to extended pelvic lymph node dissection but with a lower risk of complications \cite{SLNB_short}.
 
A drop-in gamma probe, SENSEI, has been designed to improve the accuracy of cancer or sentinel lymph node detection in RAS, as well as having other applications in cancer detection. An example of its \textit{in vivo} usage can be seen in \figureautorefname~\ref{probe_usage}. It can distinguish cancerous SLN by detecting the radiation emitted from radioactive tracers that have been injected into the body. A feasibility study has demonstrated that the drop-in gamma probe can provide accurate identification of positive nodes following the administration of technetium-99m nanocolloid  \cite{feasibility_test_3_short}.

However, relying on the live gamma level display and audible feedback from the console while the probe is scanned across the tissue surface is not an easy or intuitive way to identify radiolabeled tissue. This might impact the effectiveness of less experienced surgeons and latent hot spots may be overlooked. To address these issues, we propose a robotic scanning method to automatically and systematically examine an entire target area to locate the hot spots. In this study, we present a deep imitation training workflow based on simulation data for an end-to-end learning-based agent capable of systematically scanning target areas using visual input and the current robot state. This end-to-end approach has the potential to achieve online control without relying on pre-planning from static scene reconstruction, making it a more feasible solution for the clinical scenario where tissues are frequently deformed. The evaluation result shows that this approach is promising to automatically control the drop-in gamma probe.

\section*{MATERIALS AND METHODS}

\subsection{Building simulation environment}

\subsubsection*{Surgical scene modelling}
Surgical scene models are essential objects in the simulation. They should provide rich visual information both on texture and geometry to bring the simulated visual observation closer to reality. These were reconstructed using VisionBlender~\cite{vision_blender_short} to obtain a surface model from the SCARED dataset~\cite{SCARED_dataset_short} which contains videos of laparoscopic surgery on the da Vinci robot. Keyframes from the dataset were reconstructed and projected to a simplified mesh, resulting in a reduction of geometrical complexity while preserving a high-quality texture representation.

\begin{figure}[t]
\centering
\includegraphics[width=\columnwidth]{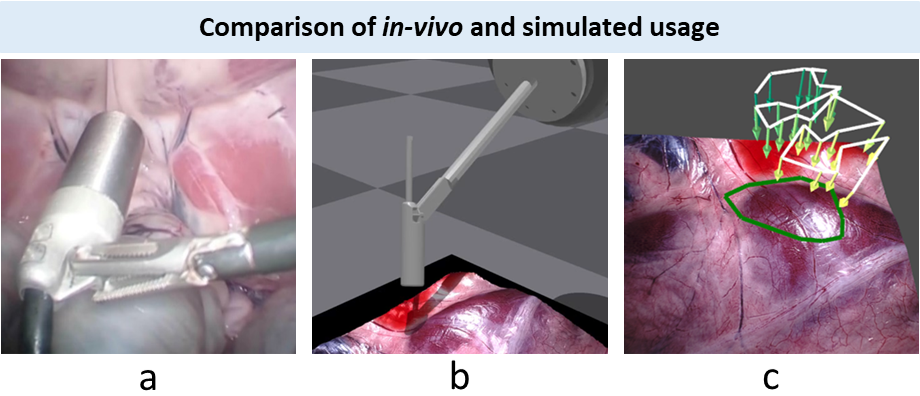}
\caption{Comparison of drop-in gamma probe usage: a) the use in an \textit{in vivo} surgical view from laparoscope \cite{huang_2020_tracking_short}; b) the  use in the simulated surgical scene c) one generated scanning path during demonstration data collection}
\vspace{-10pt}
\label{probe_usage}
\end{figure}

\subsubsection*{Robot system setup}
 The KUKA iiwa14 r820 (IIWA) robot and SENSEI drop-in gamma probe were adopted in this experiment. To mount the probe to IIWA, a customised probe holder was designed to hold the drop-in gamma probe and mount it to the flange of IIWA as shown in \figureautorefname~\ref{probe_usage} (b).
 
\begin{figure*}[t!]
\centering
\includegraphics[width=\textwidth]{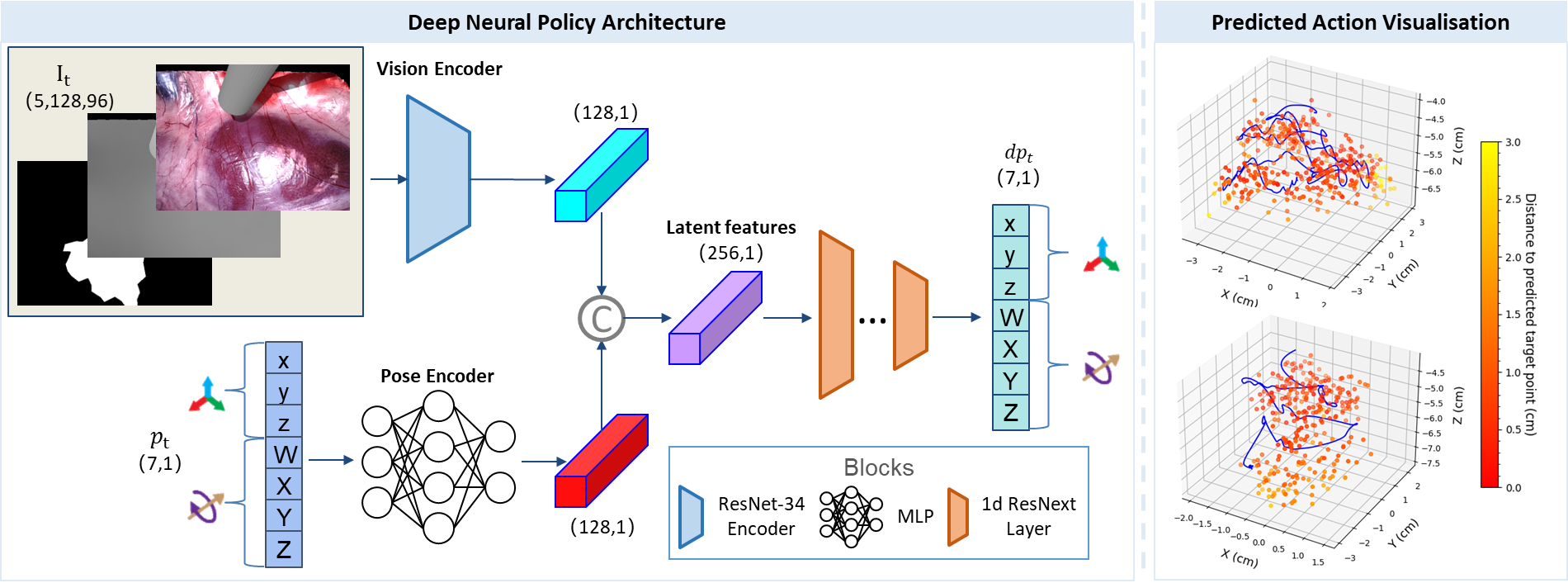}
\caption{Overview of the deep neural network policy and visualisation of predicted action. Left) The main architecture or policy network consisted of ImageNet pre-trained ResNet-34 as a vision encoder, a multilayer perceptron as a pose encoder, and 1-d ResNeXt layers as decoder to predict rotational and translational action. Right) Visualisation of predicted actions by showing the predicted next step position as the predicted action over the current position.
}
\vspace{-10pt}
\label{title_fig}
\end{figure*}

 \subsection{Demonstration data collection}
 \subsubsection*{Path generation}
 In this study, we proposed a workflow for automatic generation of large quantities of scanning demonstration data. To collect a dataset with diverse demonstrations, various target areas were selected.The waypoints were chosen to follow the standard gamma probe manipulation with an approximate distance of 3 cm from the tissue and perpendicular to the surface. Therefore, the path generation workflow contained four steps: (1) generation of a 2D target scanning area by randomly sampling points from a Gaussian distribution and fitting a convex hull to the sampled points; (2) application of a grid-based coverage path planning algorithm as a deterministic policy to generate raster scanning paths; (3) projection of the planned waypoints onto the surgical scene to obtain the 3D position and normal direction of the tissue surface; and (4) offsetting and interpolating waypoints to acquire a smooth path that satisfied the distance and pose requirements.

 \subsubsection*{Demonstration collection}
 The scanning process was simulated by Isaac Gym in the built environment to obtain photo-realistic visual observations. In the simulation, a series of targeted waypoints were defined using a generated path, and the robot smoothly navigated to these points. The camera pose was varied depending on the location and area of the target scanning area to acquire sufficient diversity. The virtual RGBD camera captured RGB images and depth maps with a binary map representing the target scanning area. The end-effector trajectories in the camera coordinates were also recorded. In total, 250 demonstrations with different target areas from five varied surgical scenes were collected.  

 \subsection{Imitation Learning}
 Imitation Learning is a supervised learning method that allows the agent to learn from expert demonstrations by observing and imitating the expert's action. In this task, the observation consisted of the current RGBD image $I_t$, a target area binary map $M$ with identical size, and the current pose $p_t=\{ x_t,y_t,z_t,W_t,X_t,Y_t,Z_t\}$ that included the Cartesian position and the orientation represented in a quaternion. The action was derived from N-step pose differences $dp_t^N = p_{t+N} - p_t$, where $N>1$ was adopted to acquire steady action. The policy network, as shown in \figureautorefname~\ref{title_fig}, utilises an encoder-decoder architecture to predict action from observations. A hybrid loss function with the weighted sum of Huber, Negative Log-likelihood, and L2 norm losses, was used to train the network.  

\subsection{Experiments}
Models were trained on an RTX 3060 graphics card for 200 epochs. The policy network was scaled up four times in experiments to maximize model performance. In the dataset, 120 demonstrations were used in training while 40 and 90 demonstrations were used for validation and evaluation. The network was evaluated by measuring the distance error on $dp_t^N(x,y,z)$ and the direct angle error on $dp_t^N(W,X,Y,Z)$.

\section*{RESULTS}
The proposed demonstration generation workflow can produce sufficient data for agent training. These demonstrations had a median trajectory length of $4.82 \, cm$ with percentiles at $2.64 \, cm (25^{th})$ and $9.12 \, cm (75^{th})$. They also had a median coverage area of $7.29 \, cm^2$ with percentiles at $5.01 \, cm (25^{th})$ and $11.22 \, cm (75^{th})$. The evaluation of the trained agent showed that the root mean square errors on each dimension of $dp_t^N(x,y,z)$ were $1.25 \, cm$, $1.29 \, cm$, and $1.24 \, cm $ with a mean distance error of $1.69 \pm 1.38 \, cm$, and an error in direct rotational angle of $14.3\pm 9.8^{\circ}$. Visualization of the predicted action points revealed a concentration around the target ground truth trajectories.

\section*{DISCUSSION}
In conclusion, the proposed training workflow could train an end-to-end vision-based gamma probe manipulation agent from the generated simulation data. The evaluation results demonstrated that it is promising to further improve the prediction accuracy of this framework and extend this method to a hardware setup. The pose error is not expected to significantly affect the gamma detection result, but may result in a larger target area on the tissue surface. Future work includes evaluating the rollout of the agent and using Reinforcement Learning methods to further explore optimal control policy.

\bibliographystyle{IEEEtran}
\bibliography{HSMR_bib}

\end{document}